\documentclass{article}

\usepackage{arxiv}

\usepackage{cite}
\usepackage{amsmath,amssymb,amsfonts}
\usepackage{algorithmic}
\usepackage{graphicx}
\usepackage{textcomp}
\usepackage{xcolor}
\usepackage[percent]{overpic}
\usepackage{paralist}
\usepackage{booktabs}
\usepackage{colortbl}
\usepackage{csquotes}
\usepackage[bookmarks=false]{hyperref}
\usepackage{fancyhdr}
\begin{document}

\title{Enhancing Resilience of Deep Learning Networks By Means of Transferable Adversaries}

\author{Moritz Seiler\\
Information Systems and Statistics \\
University of M{\"u}nster\\
M{\"u}nster, Germany \\
\texttt{moritz.seiler@uni-muenster.de}\\
\And
Heike Trautmann\\
Information Systems and Statistics \\
University of M{\"u}nster\\
M{\"u}nster, Germany \\
\texttt{trautmann@uni-muenster.de}
\And
Pascal Kerschke\\
Information Systems and Statistics \\
University of M{\"u}nster\\
M{\"u}nster, Germany \\
\texttt{kerschke@uni-muenster.de}
}

\maketitle

\thispagestyle{fancy}
\lfoot{\vspace*{-1.25cm}\rule{\columnwidth}{0.2pt}\\\footnotesize \textcopyright2020 IEEE. Personal use of this material is permitted. Permission from IEEE must be obtained for all other uses, in any current or future media, including reprinting/republishing this material for advertising or promotional purposes, creating new collective works, for resale or redistribution to servers or lists, or reuse of any copyrighted component of this work in other works.\\ This version has been accepted for publication at the \textit{International Joint Conference on Neural Networks (IJCNN)} 2020, which is part of the \textit{IEEE World Congress on Computational Intelligence (IEEE WCCI)} 2020.}\cfoot{}

\begin{abstract}
Artificial neural networks in general and deep learning networks in particular established themselves as popular and powerful machine learning algorithms. While the often tremendous sizes of these networks are beneficial when solving complex tasks, the tremendous number of parameters also causes such networks to be vulnerable to malicious behavior such as adversarial perturbations. These perturbations can change a model's classification decision. Moreover, while single-step adversaries can easily be transferred from network to network, the transfer of more powerful multi-step adversaries has -- usually -- been rather difficult.

In this work, we introduce a method for generating strong adversaries that can easily (and frequently) be transferred between different models. This method is then used to generate a large set of adversaries, based on which the effects of selected defense methods are experimentally assessed. At last, we introduce a novel, simple, yet effective approach to enhance the resilience of neural networks against adversaries and benchmark it against established defense methods. In contrast to the already existing methods, our proposed defense approach is much more efficient as it only requires a single additional forward-pass to achieve comparable performance results.
\end{abstract}


\section{Introduction}
First notions of \textit{artificial neural networks (ANNs)} -- in the context of supervised learning -- date back to the early 1940s~\cite{McCulloch1943}. Although considerable research has been conducted in the succeeding decades, people had to wait for the emergence of powerful computers, and the accompanying (significant) drop in costs for storing and processing data, until ANNs turned into highly competitive machine learning algorithms~\cite{GoodfellowBengio2016,hastie01statisticallearning}. Nowadays, ANNs represent the state of the art in a variety of supervised learning tasks, such as image, sound or object recognition \cite{HeZRS16,SimonyanZ14,AbdelHamidMJDPY14}.


However, while the strength of ANNs undoubtedly results from their huge amount of parameters, their enormous complexity also makes them rather incomprehensible. 
Although numerous works have introduced methods for interpreting the decisions made by ANNs \cite{nguyen2016synthesizing,erhan2009visualizing,zhou2016learning}, these methods lack explanations for malicious and intentional behaviour against ANNs.

\begin{figure}[t!]
\centering
    \begin{overpic}[width=0.95\linewidth, trim = 2mm 0mm 0mm 0mm, clip]{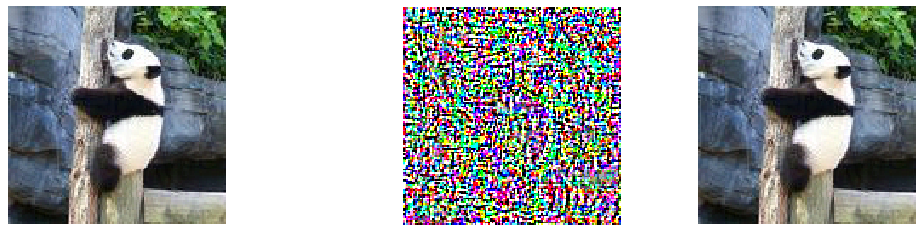}
        \put (31.5,11) {\large{$+~\varepsilon~\cdot$}}
        \put (70,11) {\large{$=$}}
    \end{overpic}
    \caption{Example for an adversary (from left to right): the original image (classified as \textit{great panda} with a confidence of 99.99\%), adversarial noise with $\varepsilon=3$ and the resulting adversarial image (classified as \textit{saxophone} with a confidence of 83.8\%).}
  \label{fig:adv_example}
\end{figure}

One possibility for fooling ANNs is the addition of carefully crafted noise, so-called \textit{adversarial perturbations} (cf. Figure~\ref{fig:adv_example}), to the model's input \cite{SzegedyZSBEGF13,goodfellow2014,GuR14,TabacofV15,FawziMF16,CarliniW16,Carlini017,KurakinGB17a}. It should be noted that these perturbations do not need to be crafted individually per network; instead, ANNs that are trained for a similar task often can be fooled by the same adversaries \cite{SzegedyZSBEGF13,goodfellow2014,PapernotMGJCS16}.

\newpage
While a plethora of works have shown the advantages of (deep learning) neural networks, their black-box characteristic makes them very vulnerable. Hence, in order to enable a more frequent and, even more importantly, more secure integration of neural networks in our daily lives, we need to ensure their robustness against adversarial attacks. 

The issue of a network's vulnerability and how to defend it against adversarial attacks will be investigated in this work. As such, our contributions can be summarized as follows:
\begin{compactenum}
    \item We present a \textbf{novel method to craft transferable adversaries}, which in turn helps to improve the understanding of a network's vulnerability against adversarial attacks.
    \item We introduce a novel approach to regularize decision boundaries and, thus, \textbf{enhance the resilience of neural networks against adversarial attacks}. Compared to previous attempts, our approach does not require additional backward-passes, which can decrease training speed significantly (cf. Table \ref{tab:training_time}).
\end{compactenum}

The remainder of this manuscript is structured as follows. In Section~\ref{sec:related} we set our work into context. We then introduce a method for crafting transferable adversaries in Section~\ref{sec:transfer}, and list possible defense strategies against adversarial attacks in Section~\ref{sec:counteractions}. Thereafter, we provide our experimental setup in Section~\ref{sec:experiments} and discuss our findings in Section~\ref{sec:results}. At last, Section~\ref{sec:conclusion} concludes our work.

\section{Related Work}
\label{sec:related}


Several methods to craft adversarial examples have been published in recent years \cite{goodfellow2014,MoosaviDezfooli15,Carlini017,madry2018towards}. In principle, one can categorize these methods into \textit{single-} and \textit{multi-step} attacks: the former only require a single gradient while the latter calculate several gradients. Kurakin et al.~\cite{KurakinGB16a} found in their work that multi-step adversaries are less transferable than single-step ones and they concluded that models are indirectly robust against multi-step adversaries. Yet, in line with the findings of \cite{dong2018boosting}, we observed that the transfer of adversaries between models is rather easy when using a so-called \emph{ensemble attack}, i.e., a multi-step attack, which is executed on an ensemble of models simultaneously \cite{dong2018boosting}.

As generating adversarial perturbations is rather simple, we will at first address the characteristics of these perturbations along with their accompanying risks of black-box attacks (see Section~\ref{sec:charact}). Note that in contrast to white-box methods, black-box approaches have no access to the model's gradient. Thereafter, in Section~\ref{sec:resil}, we briefly outline selected strategies to enhance resilience.

\subsection{Characteristics of Adversaries}
\label{sec:charact}
Szegedy et al.~\cite{SzegedyZSBEGF13} crafted artificial noise, which causes ANNs to misclassify images whose original versions they classified correctly. As a result, it could be demonstrated that already slight changes, which are hardly recognizable to the human eye, are sufficient to fool the network. As further demonstrated by \cite{SzegedyZSBEGF13,goodfellow2014}, and also in this work, adversarial perturbations are not model-specific, but instead generalize well over networks with different hyper-parameters. In addition, even networks trained on disjoint data sets, yet fulfilling a similar classification task, are likely vulnerable to the same adversarial image \cite{goodfellow2014,PapernotMGJCS16}. Thus, two models with different architectures, hyper-parameters, and trained on different data sets, often misclassify the same images. This property can be used to execute black-box attacks as demonstrated by \cite{PapernotMGJCS16}. 

Several works have been conducted to examine the effects of adversaries. The neighborhoods of adversarial examples and their original counterparts were investigated by \cite{TabacofV15}. The authors found that while already small and random noise applied to adversaries often caused it to be shifted back to their original class, the classification of the original images did not change (even when rather large random noise was applied). Therefore, the authors concluded that adversarial examples likely inhabit small sub-spaces within the decision space.

Next, \cite{tramer2017space} investigated these subspaces by measuring the number of orthogonal attack directions and found that the number of these directions is rather large. 
They concluded that the more orthogonal attack directions exist at a given point, the larger these adversarial subspaces are and the more likely adversaries transfer between models (since larger subspaces are more likely to intersect between different models).
This may be the explanation on \textit{why} adversarial perturbations are often universal across different model designs and often universal across datasets.

\subsection{Enhancing Resilience through Adversarial Training}
\label{sec:counteractions}
\begin{figure}[t!]
\flushright
    \begin{overpic}[width=0.92\columnwidth]{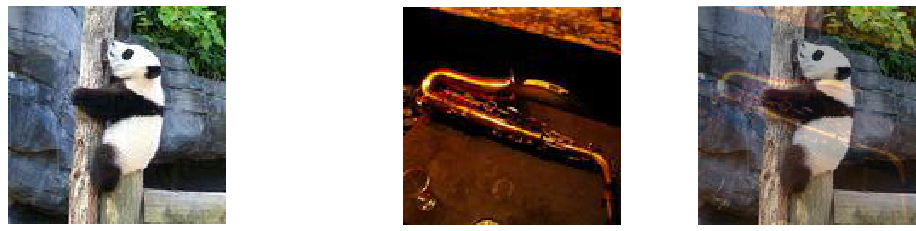}
        \put (-8,11) {{$(1-\alpha)~\cdot$}}
        \put (29.5,11) {{$+~\alpha~\cdot$}}
        \put (70,11) {{$=$}}
    \end{overpic}
	\caption{Idea of our proposed method: linear combination of the original image (\textit{great panda}) with a randomly selected image (\textit{saxophone}) using weight $\alpha \in (0, 0.5)$ (here: $\alpha = 0.3$). The resulting training image (right) inherits the class of the original image. This allows us to generate, on the one hand, more training examples and, on the other one, training examples which lie in close proximity to the decision boundaries.}
  \label{fig:alpha_example}
\end{figure}
\label{sec:resil}

Defense strategies against adversarial attacks can mainly be distinguished into two categories: approaches that either change (i) a model's topology, or (ii) the way a model is trained. Gu and Rigazio~\cite{GuR14} stated that a model's robustness \enquote{is more related to intrinsic deficiencies in the training procedure and objective function than to [its] topology}.



Under the concept of \textit{Adversarial Training}, several works have introduced regularizers which decrease a model's vulnerability to adversarial perturbations. The first one was introduced by Goodfellow et al.~\cite{goodfellow2014}. Here, the objective function of the \textit{Fast Gradient Sign Method} (FGSM) \cite{goodfellow2014} is added to the classification loss. 
For every training image, a single-step adversary is calculated and the network's sensitivity to this adversary is minimized. Thus, \textit{Adversarial Training} acts as a regularizer to improve resilience \cite{goodfellow2014,KurakinGB16a}.


The training method of \cite{goodfellow2014} only uses a single back-propagation step to calculate adversarial perturbations. In contrast to the aforementioned approach, other works introduced a min-max optimization method \cite{Huang2015LearningWA,shaham2015}. It prepares the network for the worst-case scenario under a limited perturbation and minimizes its sensitivity to it. The algorithm works in two steps. First, it maximizes the loss of the input image by adding a small adversarial perturbation. However, the network trained with a min-max regularizer performed worse on the classification tasks \cite{Huang2015LearningWA} -- even when trained with rather small perturbations. Due to the additional maximization problem, the algorithm is even more complex to execute, which further increases training time.

Similar approaches were introduced by \cite{Miyato2015DistributionalSW,madry2018towards}. The approach of \cite{Miyato2015DistributionalSW}, denoted \textit{Virtual Adversarial Training} (VAT), works in a similar way as the min-max optimization, yet it maximizes the Kullback-Leibler divergence \cite{Kullback51klDivergence} between the probability vectors of the training image and an adversary. The KL-divergence between the original and an adversary can be interpreted as the local smoothness of the decision space. Thus, the network not only learns to classify the training images correctly, but also smoothens the decision boundaries around each of the training points. Miyato et al.~\cite{Miyato2015DistributionalSW} used \textit{Projected Gradient Descent} (PGD) to solve the inner maximization problem. The authors could demonstrate a guaranteed robustness even against multi-step adversaries under small $l_\infty$-norm -- with $l_\infty = max\{|x~-~x^{adv}|\}$ -- where $x$ is the unchanged input image while $x^{adv}$ is the image with additional adversarial perturbation. Multi-step black- and white-box attacks have been conducted under a $l_\infty\leq8$-constraint. Even under multi-step white-box attacks, the accuracy of their model is 45\%, whereas the accuracy of a non-robustified, and thus defenseless, model would be close to zero. However, as there are additional forward-backward passes required, training time takes accordingly longer.

\section{Crafting Strong and Transferable Adversaries}
\label{sec:transfer}
\begin{figure}[t!]
\centering
	\includegraphics[width=\columnwidth, trim = 0mm 8mm 0mm 0mm, clip]{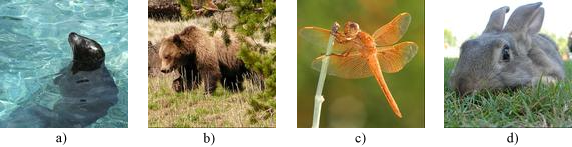}
	\caption[Random Images from the Dataset]{Exemplary images from the modified \textit{Large Scale Visual Recognition Challenge 2017} dataset: Seal, Bear, Dragonfly and Rabbit (left to right).}
	\label{fig:example_pic}
\end{figure}

Goodfellow et al.~\cite{goodfellow2014} introduced the \textit{Fast Gradient Sign Method} (FGSM):
\begin{align}
    x^{adv} &= x + \varepsilon \cdot \text{sign} \left(\nabla_{x}~L(\Theta, x, y)\right) \label{eq:fgsm}
\end{align}
The loss function\footnote{In this work, we refer to the cross-entropy as loss function. Yet, any other function can most likely be used, too.} $L$ of the model $\Theta$ is differentiated with respect to the input image $x$ and its true class $y$. The sign of the resulting gradient $\nabla_{x}~L$ is used to calculate adversarial noise which, when applied to an image, increases the model's loss and, thereby, could shift its classification decision. The parameter $\varepsilon$ is used to control the amount of perturbations. Note that a larger $\varepsilon$ does not necessarily lead to a higher chance of fooling a model as demonstrated by Figure \ref{fig:loss}. Instead, tuning $\varepsilon$ is highly important. The resulting adversary is a single-step one.

One can execute the FGSM multiple times with a small $\varepsilon$-value to converge towards the adversarial sub-space iteratively. The resulting adversaries are then referred to as multi-step ones. In order to find multi-step adversaries under limited amount of perturbations and close to the original image, several methods such as \cite{MoosaviDezfooli15,Carlini017,madry2018towards,KurakinGB16a} have been introduced. When the perturbation exceeds a certain limit -- usually the $l_\infty$- or $l_2$-norm of the adversarial perturbation is used -- the perturbation is projected back into the search space, e.g., by using pixel-wise clipping.

As pointed out by \cite{TabacofV15,tramer2017space}, adversaries lie in sub-spaces which are in close proximity to true training examples and which often intersect between models. One can argue that when crafting adversaries with multi-step methods, these adversaries often do not lie in intersecting sub-spaces because the resulting adversary may be overfitted to the particular model as its loss is maximized. To overcome this issue, \cite{dong2018boosting} introduced the (i) \textit{Momentum Iterative Gradient Based Method} (MI-FGSM) and (ii) the \textit{ensemble in logits}. The former method uses a momentum to overcome local maxima and the latter one combined the logits of an ensemble of models to find a common and intrinsic direction. Attacking an ensemble of models increases the likelihood of transferable adversaries as these adversaries are not optimized for a single model but for several models simultaneously. Hence, the adversaries lie more likely in intersecting adversarial sub-spaces. 

In contrast to \cite{dong2018boosting}, we used the combined gradient of an ensemble of models to craft adversaries which lie in intersecting sub-spaces. We call our method \textit{gradient ensemble attack}. We found that these adversaries are likely to transfer between models of the same architecture -- and frequently transfer to other architectures as well (see Section~\ref{sec:results}). To ensure that every model's gradient has the same impact on the resulting adversarial perturbation, we normalize each gradient to length one:
\begin{align}
    \hat{\nabla}_{x}~L(\Theta, x, y) &= \frac{\nabla_{x}~L(\Theta, x, y)}{||\nabla_{x}~L(\Theta, x, y)||_2}. \label{eq:grad_norm}
\end{align}
Afterwards, the different gradients are summed up to find a common direction of intersecting adversarial sub-spaces. Similar to \cite{KurakinGB16a}, we approximated the true direction by using the sign-method of the summed and normalized gradients to the image in an iterative process:
\begin{eqnarray}
 x^{adv}_{i} &=& x_{i-1} + \lambda \cdot \text{sign} \left(\sum_{n=0}^N~\hat{\nabla}_{x}~L(\Theta_n, x_{i-1}, y) \right) \label{eq:adv_methode}\\
  \text{s.t.} && ||x^{adv}_{i} - x||_{\infty}~\leq~\varepsilon \quad \text{and} \quad x_0 = x.\nonumber
\end{eqnarray}
To ensure that the magnitude of the perturbations stays within a given limit, we used pixel-wise clipping. Further, we used a learning rate of $\lambda=1$ (to slowly convert towards the adversarial spot) and set the number of steps to $I=\min(\varepsilon + 4, 1.25\cdot\varepsilon)$ as proposed by \cite{KurakinGB16a} to craft adversarial images. In addition, we chose $\varepsilon\in\{4,8,16\}$ to limit the amount of perturbations in order to test and compare different magnitudes of adversarial perturbations.

\section{Proposed Defense Strategies}
As demonstrated by \cite{TramerKPGBM18} and \cite{athalye18a}, adversarial training may lead to \emph{gradient masking}. This term describes a phenomenon in which the decision boundaries are roughly the same, yet the corresponding gradient is damaged or obfuscated in a way that white-box attacks fail. However, the model does not become more resilient against adversarial examples in black-box attack scenarios or against transferable adversaries as these attacks are based on surrogate models.

In order to avoid the risk of gradient masking, we propose a method that does not require gradient calculations and still flattens the decision space. In addition, we avoid the expensive optimization of an inner maximization problem as done in VAT or PGD. We found that while random noise is mostly ignored by the model, superimposing two pictures does distract a model. Therefore, as illustrated in Figure~\ref{fig:alpha_example}, we designed training images $\tilde{x}_i$ by placing a randomly selected image $x_r$ on top of the original training image $x_i$:
\begin{align}
    \tilde{x}_i = (1-\alpha) \cdot x_i + \alpha \cdot x_r. \label{eq:alpha_method}
\end{align}
The parameter $\alpha \in (0, 0.5)$ controls the impact of image $x_r$ on the generated image $\tilde{x}_i$. As the majority of the image originates from the original image $x_i$, the generated image $\tilde{x}_i$ will inherit the class label $y_i$ of the original image. Our proposed approach thus allows to (i) generate \emph{more training examples}, and (ii) create images that are closer to the decision boundaries of at least two classes -- and thus \emph{harder to distinguish} -- as $\tilde{x}_i$ contains properties of two image classes. Thereby, the space between the different classes is flattened and the boundaries are regularized. Thereafter, the networks will be trained by minimizing the loss $L(\Theta,\tilde{x}_i, y_i)$.
As the results depend on the choice of $\alpha$, we considered three different approaches:
\begin{compactenum}
\item Using a fixed $\alpha$-value.
\item Following the idea of \cite{Miyato2015DistributionalSW}, i.e., first predicting the classes $y_i$ and $\tilde{y}_i$ based on $x_i$ and $\tilde{x}_i$ (and using a fixed $\alpha$ as in 1.), and then minimizing the loss of the unmodified training examples \emph{and} the KL-Divergence between the predicted classes of the unmodified and modified training examples:
\[\underset{\Theta}{\text{minimize}} \quad L(\Theta, x_i, y_i) + \lambda \cdot KL(\hat{y}_i~||~\tilde{y}_i).\]
\item Similarly to 2., but this time drawing $\alpha$ from a beta-distribution instead of a fixed $\alpha$-value; $\alpha \sim B(p,q)$ with $p=2$ and $q \in \{4,6,10\}$.
\end{compactenum}
Note that while the first method does not require any additional passes, the latter two methods require a second forward-pass to calculate the KL-Divergence -- but no additional backward-pass. As stated by \cite{Miyato2015DistributionalSW}, the KL-divergence can be interpreted as the local smoothness of the decision space. If the divergence is high, the predictions of both input images are dissimilar, implying that the decision space between the two activations is not flattened. By minimizing the divergence, the models learn to find similar activations for both images and, thereby, flatten the decision space between the two activations.

As a side effect, our method allows to generate a multitude of training images (using Equation~(\ref{eq:alpha_method})), which in turn allows to train a larger area.
\begin{figure}[t!]
\centering
	\includegraphics[width=0.95\columnwidth]{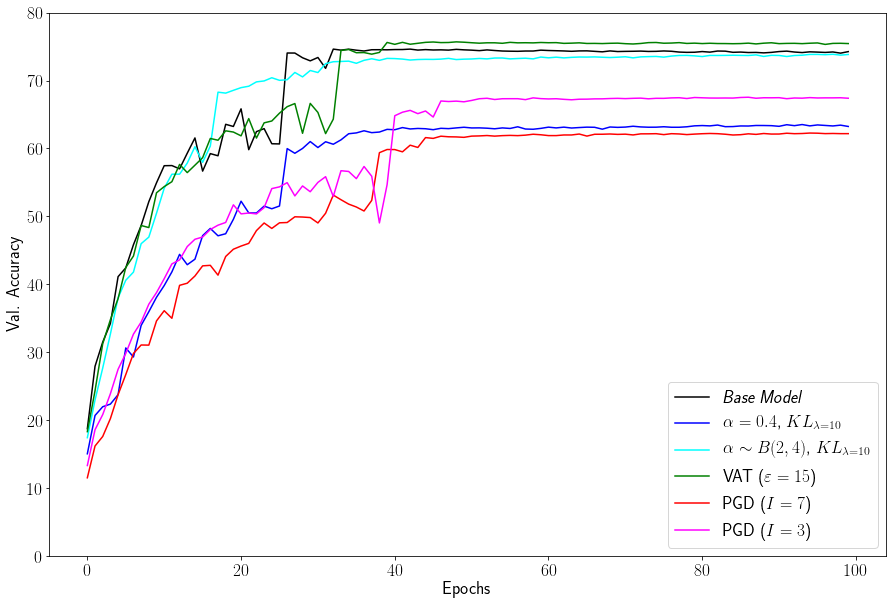}
	\caption[Validation Accuracy during training]{Visualization of the validation accuracy during training. The \textit{Projected Gradient Descent} \cite{madry2018towards} substantially reduces the overall accuracy compared to the \textit{Base Model}, our proposed method, or the \textit{Virtual Adversarial Training} \cite{Miyato2015DistributionalSW}. Furthermore, the model $\alpha \sim B(2, 4)$, $KL_{\lambda=10}$ achieves an accuracy of more than 70\% with fewer methods than the \textit{Base Model}.}
	\label{fig:acc_druing_training}
\end{figure}

\section{Methodology and Experiments}
\label{sec:experiments}
In the following, we briefly outline the models' topologies and used training data.

\subsection{Dataset}
Within our experiments, we used the data from the second challenge of the \textit{Large Scale Visual Recognition Challenge 2017}\footnote{\url{http://image-net.org/challenges/LSVRC/2017/}} \cite{ILSVRC15} for training and validating our models. 
Yet, as down-scaling the whole images to the required size was not reasonable, we first used bounding boxes to cut out the objects within each of the images. Then, as the bounding boxes were of different sizes, the cropped images were down-scaled to the desired resolution of 128 $\times$ 128 pixels (cf. Figure \ref{fig:example_pic}).

As a result, our dataset contains roughly 400,000 images. Unfortunately, the images are not uniformly distributed across all 200 classes. As the mismatch between the classes is too large to use oversampling, we selected 100 classes with an identical number of images per class. For training and validation purposes, we performed holdout with a 80-20-split. Thereby, we created a dataset which is larger and has a higher resolution than CIFAR10/100~\cite{krizhevsky2009learning} while (at the same time) being significantly smaller and less complex than ImageNet~\cite{ILSVRC15}. In addition, by using balanced classes, we eliminate any side-effects which may occur from unbalanced datasets and may influence our results.

\subsection{Models}
\begin{table}[!t]
    \centering
    \caption{Measured training Time for different models (all models were trained on a single Nvidia Quadro RTX 6000). Training a single epoch with our proposed method takes only slightly longer than the \textit{Base Model} without any defense. In addition, $\alpha \sim B(2, 4)$, $KL_{\lambda=10}$ reaches 60\% and 70\% Validation Accuracy about 30 minutes earlier than the \textit{Base Model} as it converges faster (cf. Figure \ref{fig:acc_druing_training}). Training models using the \textit{Projected Gradient Descent} \cite{madry2018towards} or \textit{Virtual Adversarial Training} \cite{Miyato2015DistributionalSW} takes noticeably longer per epoch, and they take even longer to converge.}
    \label{tab:training_time}
        \begin{tabular}{l|cccc}
                       & Median Time & Time until & Time until & Time until \\
            Model      & p. Epoch    & 50\% Acc.  & 60\% Acc.  & 70\% Acc. \\
            \midrule
            \textit{Base} &  \cellcolor{gray!25}\textbf{6:43m} & \cellcolor{gray!25}\textbf{00:59h} & 02:15h & 03:17h \\
            $\alpha=0.4$   & 6:55m & 02:18h & 03:14h & -/- \\
            $\alpha \sim B(2, 4)$ & 6:51m & 01:01h & \cellcolor{gray!25}\textbf{01:42h} & \cellcolor{gray!25}\textbf{02:44h} \\
            VAT ($\varepsilon=15$) & 14:45m & 02:12h & 03:55h & 08:05h \\
            PGD ($I=7$) & 22:40m & 11:46h & 15:56h & -/- \\
            PGD ($I=3$) & 11:37m & 03:53h & 07:45h & -/- 
        \end{tabular}
\end{table}
To investigate the effects of adversarial perturbations, we trained multiple ANNs. First, an ``unprotected'' model -- denoted \textit{base model} in the following -- is trained. For this purpose, we took a VGGNet~\cite{SimonyanZ14}, as it provides a straightforward and uncluttered design of convolutions, pooling and dense layers. In addition, we modified the design by using a \textit{Global-Average-Pooling} layer \cite{LinCY13} and extended each CNN layer with a \textit{Batch Normalization} layer \cite{Ioffe15}. Afterwards, we compared our proposed methods to different ResNets \cite{HeZRS16} to verify our findings.

In order to compare different methods to defend against adversarial attacks, we trained several models with and without different defense methods (see Section~\ref{sec:results} for more details). All models were trained on a single Nvidia RTX 6000 or a single Nvidia V100 GPU. We found, that by using VAT \cite{Miyato2015DistributionalSW} or PGD \cite{madry2018towards} the training speed is reduced, significantly. 
Each training epoch not only takes longer, but the models also converge more slowly towards the optimum. 
In contrast, our proposed method is more time efficient as it only requires an additional forward pass and it even converges faster (cf. Figure \ref{fig:acc_druing_training} and Table \ref{tab:training_time}).

\subsection{Assessing a Network's Resilience against Adversarial Perturbations}
\label{sec:assessment}
To assess the robustness of a network against adversaries, we trained six models per considered network architecture. We then used an ensemble of one, three or five of the six ANNs\footnote{Note that our method applied to an ensemble of one model is identical to the i-FGSM.} (see Section~\ref{sec:results} for details) to extract a set of adversarial images using Equation~(\ref{eq:adv_methode}). More precisely, an image is considered being adversarial, if it is misclassified by \emph{all} of the ensemble's networks -- note that for simplicity the networks do not have to agree on the same wrong class. The set of the extracted adversarial images is then classified by the sixth model and its accuracy is taken as quality indicator of the respective network architecture.

\section{Results and Discussion}
\label{sec:results}
\begin{table*}[!t]
    \centering
    \caption{Accuracy of different networks on the sets of original and adversarial images. All adversarial datasets are crafted using our proposed \textit{gradient ensemble attack} approach with one, three or five (vulnerable) VGGNet13 models. Our method is equivalent to \cite{KurakinGB16a,madry2018towards} if only one model is used. The shown results are for $\varepsilon=16$. When training a model with PGD, the performance on the original images is significantly lower in comparison to the \textit{Base Model}. In contrast, VAT and our proposed method slightly increase the validation accuracy. Across all adversarial perturbations, VAT models achieved the highest accuracy. Noteworthy values are highlighted.
    }
    \label{tab:results}
        \begin{tabular}{r|cc|cc|cc|cc}
        &\multicolumn{2}{c|}{\bf Accuracy on} & \multicolumn{6}{c}{\bf Accuracy on Adversarial Dataset based on} \\
        &\multicolumn{2}{c|}{\bf Original Data} & \multicolumn{2}{c|}{\bf One Model} & \multicolumn{2}{c|}{\bf Three Models} & \multicolumn{2}{c}{\bf Five Models} \\
        Parametrization \, & \, Train &Val. \, & \, True& Adv. \, & \, True \, & \, Adv. \, & \, True \, & \, Adv.\\
        \midrule
        \multicolumn{9}{l}{\textit{Base Model}}\\
         \, & \, 91.2\% \, & \, \cellcolor{gray!25}\textbf{73.4}\% \, & \, 89.9\% \, & \, \cellcolor{gray!25}\textbf{23.8}\% \, & \, 95.3\% \, & \, \cellcolor{gray!25}\textbf{4.1}\% \, & \, 96.7\% \, & \, \cellcolor{gray!25}\textbf{0.9}\% \\
        \midrule
        \multicolumn{9}{l}{\textit{Projected Gradient Descent (PGD) with} $\varepsilon = 8$ \textit{and} $\lambda = 2$}\\
        $I = 3$ \, & \, 86.7\% \, & \, \cellcolor{gray!25}\textbf{66.2}\% \, & \, 79.6\% \, & \, 79.0\% \, & \, 84.6\% \, & \, 83.6\% \, & \, 86.6\% \, & \, 86.5\% \\
        $I = 7$ \, & \, 77.8\% \, & \, \cellcolor{gray!25}\textbf{62.4}\% \, & \, 75.6\% \, & \, 75.1\% \, & \, 80.7\% \, & \, 80.0\% \, & \, 82.8\% \, & \, 81.8\% \\
        \midrule
        \multicolumn{9}{l}{\textit{Virtual Adversarial Training (VAT) with} $I = 3$ \textit{and} $\lambda = 1$}\\
        $\varepsilon = \phantom{1}5$ \, & \, 97.7\% \, & \, 76.2\% \, & \, 90.6\% \, & \, 85.1\% \, & \, 95.2\% \, & \, \cellcolor{gray!25}\textbf{87.2}\% \, & \, 96.6\% \, & \, 86.9\% \\
        $\varepsilon = 15$ \, & \, 99.0\% \, & \, 74.6\% \, & \, 88.5\% \, & \, \cellcolor{gray!25}\textbf{86.0}\% \, & \, 93.0\% \, & \, 86.0\% \, & \, 94.5\% \, & \, \cellcolor{gray!25}\textbf{90.3}\% \\
        $\varepsilon = 25$ \, & \, 96.9\% \, & \, 76.5\% \, & \, 91.1\% \, & \, 83.1\% \, & \, 95.5\% \, & \, 82.8\% \, & \, 96.8\% \, & \, 80.2\% \\
        \midrule
        \multicolumn{9}{l}{\textit{Our Proposed Method} with $KL_{\lambda=10}$}\\
        $\alpha \sim B(2, \phantom{1}4)$ \, & \, 90.6\% \, & \, 74.3\% \, & \, 88.8\% \, & \, 71.9\% \, & \, 93.5\% \, & \, 62.1\% \, & \, 95.1\% \, & \, 53.7\% \\
        $\alpha \sim B(2, \phantom{1}6)$ \, & \, 84.7\% \, & \, 72.4\% \, & \, 87.1\% \, & \, 74.2\% \, & \, 91.8\% \, & \, 70.0\% \, & \, 93.7\% \, & \, 66.0\% \\
        $\alpha \sim B(2, 10)$ \, & \, 86.9\% \, & \, 71.4\% \, & \, 85.8\% \, & \, 76.9\% \, & \, 90.4\% \, & \, 75.0\% \, & \, 92.4\% \, & \, 72.4\% \\
        \end{tabular}
\end{table*}

At first, adversaries have been generated as described in Section~\ref{sec:assessment}. For the gradient alignment, we considered an ensemble of one, three and five models, respectively. In a first analysis, we investigated the impact of the perturbation parameter for the values $\varepsilon \in\{4,8,16\}$. Interestingly, we observed that the success rate for crafting adversaries is not sensitive to the tested values for $\varepsilon$. Therefore, we are only referring to the adversaries with $\varepsilon=16$ as they are the most powerful ones. If a single unprotected model -- which is fully identical to the \textit{Base Model} in terms of topology and training execution -- is used to calculate multi-step adversaries, the \textit{Base Model's} classification accuracy is still $23.8\%$ as shown in Table~\ref{tab:results}. However, aligning the gradients of an ensemble of three or five models, the \textit{Base Model's} accuracy on these adversaries decreases to $4\%$ and $0.9\%$, respectively.

Next, we trained multiple models with \textit{Virtual Adversarial Training (VAT)}, \textit{Projected Gradient Descent Training (PGD)} and the three different defense methods proposed in this work (see Section~\ref{sec:counteractions}). For VAT we used $I=3$, $\lambda=1$ and $\varepsilon\in\{5,10,15,20,25\}$. Miyato et al. \cite{Miyato2015DistributionalSW} recommended 
using $I=1$ and $\lambda=1$ as they found it to be sufficient. We increased the \textit{power of iterations} to $I=3$ to ensure a better conversion (cf. Miyato et al. \cite{Miyato2015DistributionalSW} for more details). As tuning $\varepsilon$ is most important, we tried several different values and compared them to each other. Next, we adapted the default parameter for PGD as proposed by \cite{madry2018towards}: $\varepsilon=8$, $\lambda=2$ and $I\in\{3, 7\}$ as the number of iterations. $\varepsilon=8$, $\lambda=2$ and $I=\{7\}$ are the default settings used for on CIFAR10 by \cite{madry2018towards}. In addition, we used $I=3$ to speed up training.

Table \ref{tab:results} shows the results of different methods based on their accuracy on our crafted adversarial images. As indicated by the base model's accuracy values on the adversarial data, the more models are used for our proposed \textit{gradient ensemble attack}, the higher is the success rate of transferring the adversaries to other models. This demonstrates that our adversaries, crafted from an ensemble of models, are likely transferable to other networks. Moreover, the VAT models seem to perform best on adversarial images.

To test the generalizability of our approach, we additionally assessed our adversaries on ResNet32 and Res\-Net50~\cite{HeZRS16}. As shown in Table~\ref{tab:transfer}, when applying a \textit{gradient ensemble attack} on VGGNet13 and the ResNet models together, the resulting adversaries likely transfer between both topologies. The accuracy of unprotected (base) models on our combined adversarial dataset is 0.01\% for the VGGNet13 network and about 26 to 27\% for both ResNet models -- although all three models have an accuracy of over 90\% on the original images. Even adversaries that were originally crafted for a different topology reduce a model's accuracy noticeably. 
    
We further tested our method on different ResNets. As shown in the bottom half of Table~\ref{tab:transfer}, we found that not only the originally considered VGGNet13 models, but also ResNet32 and ResNet50 became more resilient against the transferable adversaries.

However, comparing the performance of adversarial defense methods merely based on the model's accuracy or on the success rate for crafting adversaries is problematic. Adversarial sub-spaces may occur a little aside of the original ones or gradient masking could prevent gradient-based attack methods from being successful. Therefore, we do not only refer to a model's accuracy on strong and transferable adversaries, but also investigate the surrounding decision space, as well as its gradient.

\begin{figure*}[p]
    \centering
    \centering
    \begin{minipage}[b]{0.325\textwidth}
        \begin{overpic}[width=\textwidth]{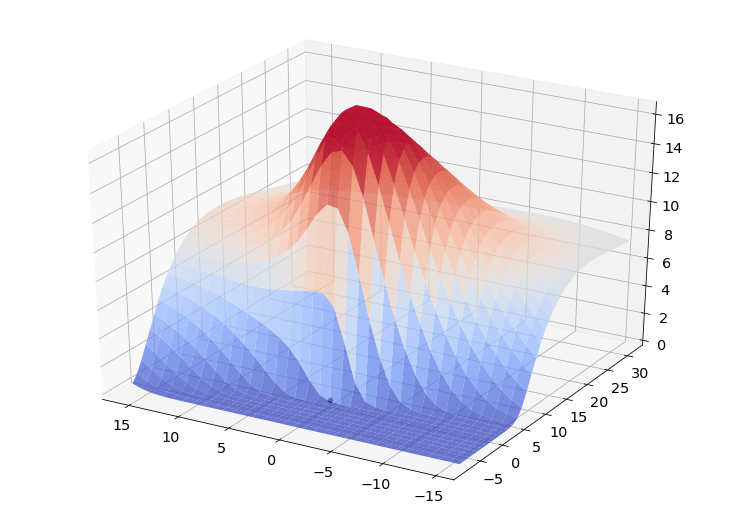}
                \put (40,20) {\tiny{$X$}}
                \put (30,2) {$\varepsilon_1$}
                \put (80,5) {$\varepsilon_2$}
            \end{overpic}\\
            (a) \textit{Base Model}
    \end{minipage}
    \hfill
    \begin{minipage}[b]{0.325\textwidth}
        \begin{overpic}[width=\textwidth]{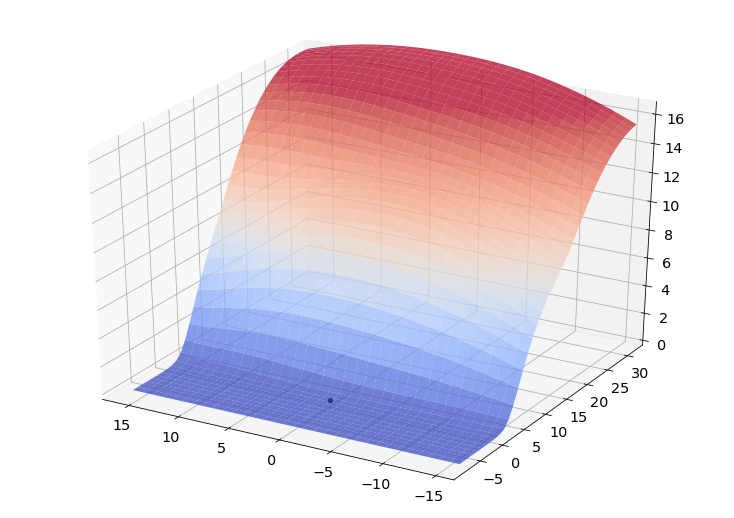}
                \put (40,20) {\tiny{$X$}}
                \put (30,2) {$\varepsilon_1$}
                \put (80,5) {$\varepsilon_2$}
            \end{overpic}\\
            (b) VAT ($\varepsilon=15$) \cite{Miyato2015DistributionalSW}
    \end{minipage}
    \hfill
    \begin{minipage}[b]{0.325\textwidth}
        \begin{overpic}[width=\textwidth]{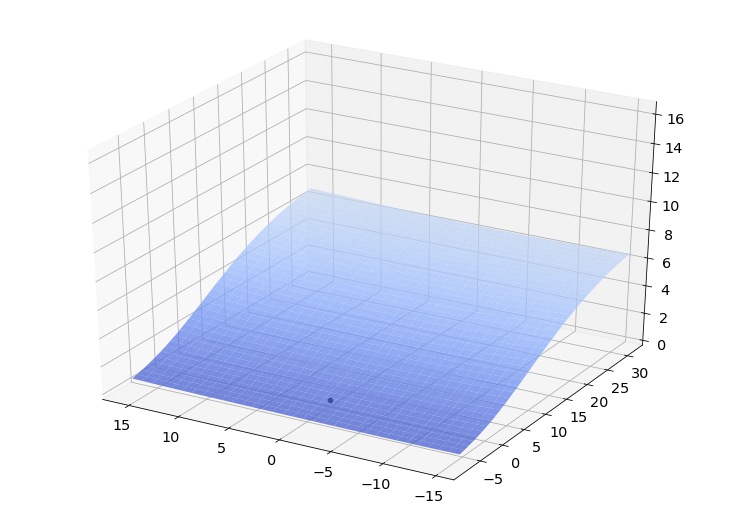}
                \put (40,20) {\tiny{$X$}}
                \put (30,2) {$\varepsilon_1$}
                \put (80,5) {$\varepsilon_2$}
            \end{overpic}\\
            (c) PGD ($I=7$) \cite{madry2018towards}
    \end{minipage}\\
    \medskip
    \begin{minipage}[b]{0.325\textwidth}
        \begin{overpic}[width=\textwidth]{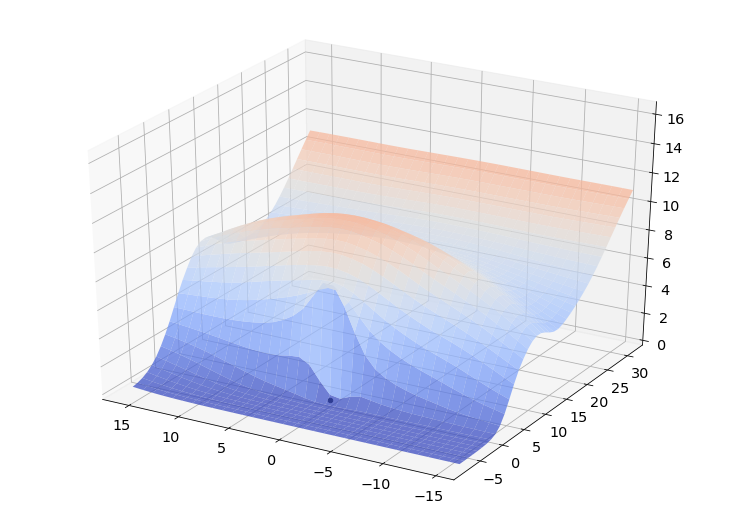}
                \put (40,20) {\tiny{$X$}}
                \put (30,2) {$\varepsilon_1$}
                \put (80,5) {$\varepsilon_2$}
            \end{overpic}\\
            (d) $\alpha=0.4$
    \end{minipage}
    \hfill
    \begin{minipage}[b]{0.325\textwidth}
        \begin{overpic}[width=\textwidth]{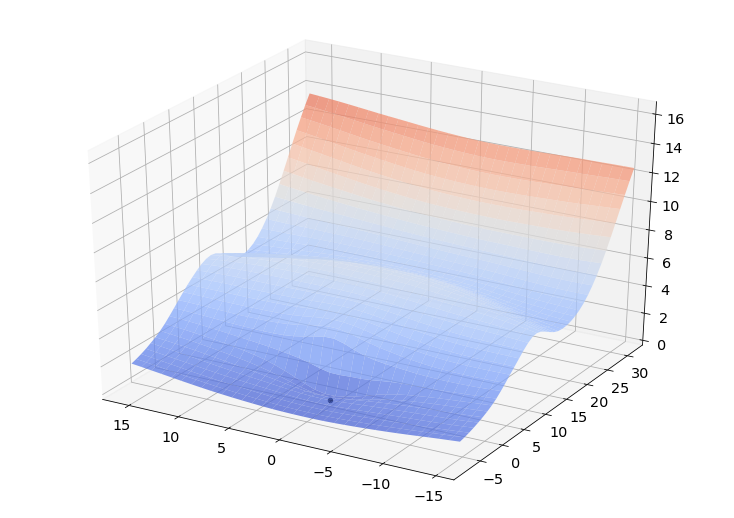}
                \put (40,20) {\tiny{$X$}}
                \put (30,2) {$\varepsilon_1$}
                \put (80,5) {$\varepsilon_2$}
            \end{overpic}\\
            (e) $\alpha=0.4$, $KL_{\lambda=10}$
    \end{minipage}
    \hfill
    \begin{minipage}[b]{0.325\textwidth}
        \begin{overpic}[width=\textwidth]{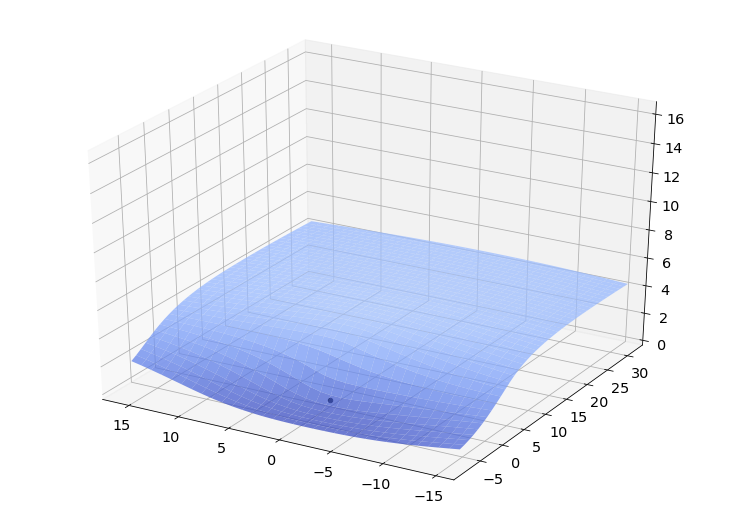}
                \put (40,20) {\tiny{$X$}}
                \put (30,2) {$\varepsilon_1$}
                \put (80,5) {$\varepsilon_2$}
            \end{overpic}\\
            (f) $\alpha \sim B(2, 4)$, $KL_{\lambda=10}$
    \end{minipage}
    \caption{Visualization of the decision space of six different VGGNet13 models in two adversarial directions of the same input image $X$. The loss is plotted using $x^* = x + \varepsilon_1 \cdot \text{sign} \left(\nabla_{x}~L_1\right) + \varepsilon_2 \cdot \text{sign} \left(\nabla_{x}~L_2\right)$ (cf. \cite{TramerKPGBM18}). The red peak is an adversarial spot, which corresponds to a very high loss. The magnitude of the related model's gradients $\text{sign}\left(\nabla_{x}~L_2\right)$ is controlled by $\varepsilon_2\in[-8,32]$, whereas $\varepsilon_1\in[-16,16]$ controls the magnitude of an unprotected model's gradient $\text{sign}\left(\nabla_{x}~L_1\right)$. The six images visualize how the loss values change when the input $x$ is moved in one of these two directions. The images in (d) to (f) correspond to our proposed models.}
    \label{fig:gradient}
    
    \bigskip
        \begin{minipage}[b]{0.325\textwidth}
            \begin{overpic}[width=\textwidth]{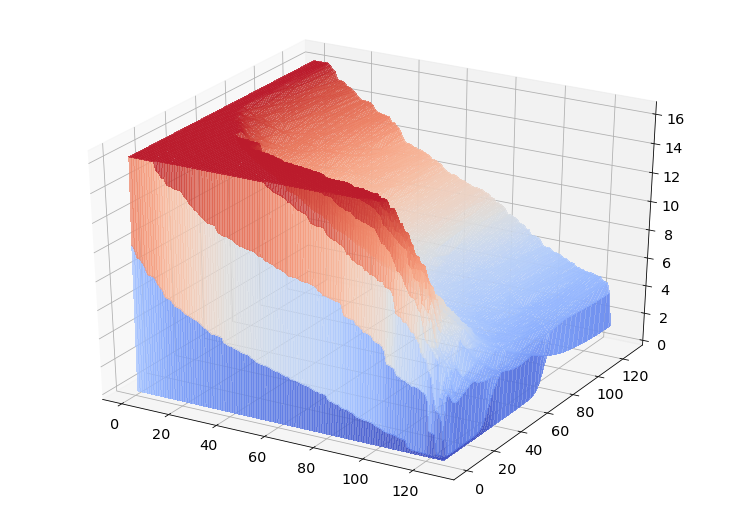}
                    \put (30,1) {$X$}
                    \put (80,5) {$\varepsilon$}
                \end{overpic}\\
                (a) \textit{Base Model}
        \end{minipage}
        \hfill
        \begin{minipage}[b]{0.325\textwidth}
            \begin{overpic}[width=\textwidth]{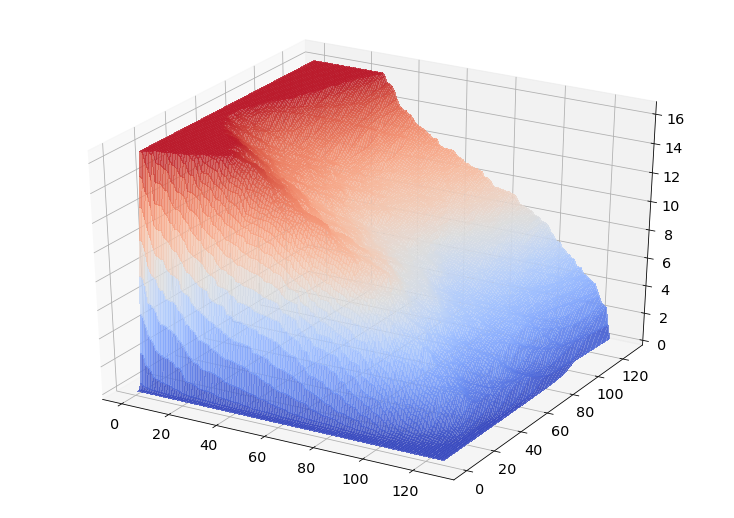}
                    \put (30,1) {$X$}
                    \put (80,5) {$\varepsilon$}
                \end{overpic}\\
                (b) VAT ($\varepsilon=15$) \cite{Miyato2015DistributionalSW}
        \end{minipage}
        \hfill
        \begin{minipage}[b]{0.325\textwidth}
            \begin{overpic}[width=\textwidth]{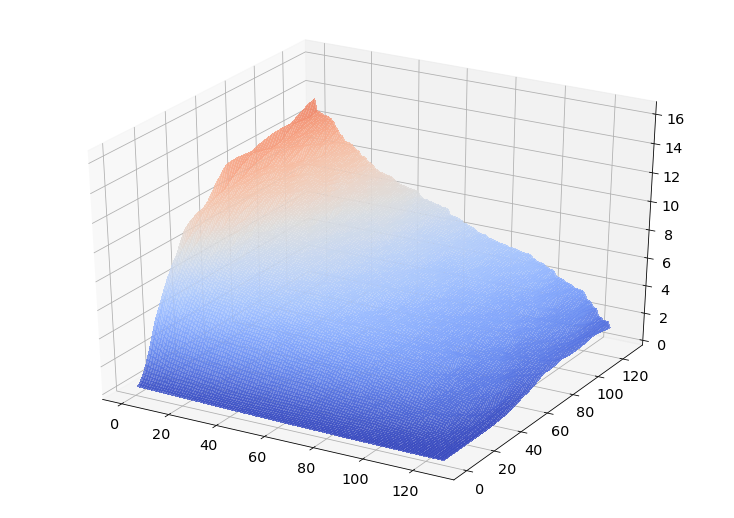}
                    \put (30,1) {$X$}
                    \put (80,5) {$\varepsilon$}
                \end{overpic}\\
                (c) PGD ($I=7$) \cite{madry2018towards}
        \end{minipage}\\
        \medskip
        \begin{minipage}[b]{0.325\textwidth}
            \begin{overpic}[width=\textwidth]{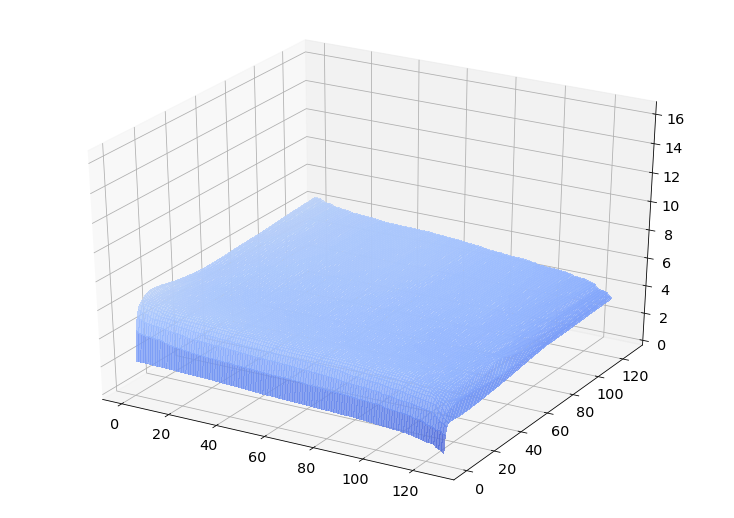}
                    \put (30,1) {$X$}
                    \put (80,5) {$\varepsilon$}
                \end{overpic}\\
                (d) VGGNet13 (Ours)
        \end{minipage}
        \hfill
        \begin{minipage}[b]{0.325\textwidth}
            \begin{overpic}[width=\textwidth]{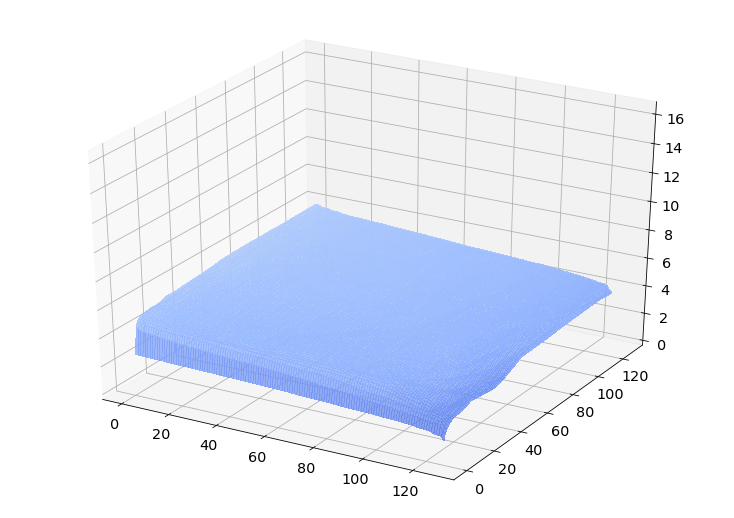}
                    \put (30,1) {$X$}
                    \put (80,5) {$\varepsilon$}
                \end{overpic}\\
                (e) ResNet32 (Ours)
        \end{minipage}
        \hfill
        \begin{minipage}[b]{0.325\textwidth}
            \begin{overpic}[width=\textwidth]{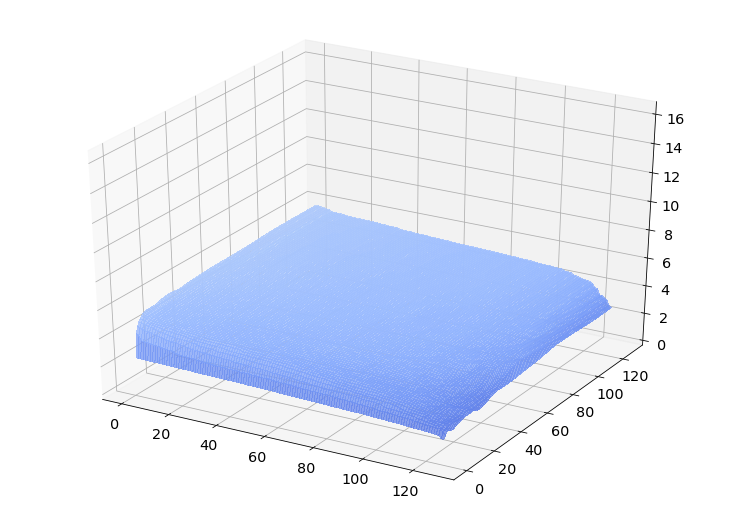}
                    \put (30,1) {$X$}
                    \put (80,5) {$\varepsilon$}
                \end{overpic}\\
                (f) ResNet50 (Ours)
        \end{minipage}
        \caption{Impact of different images $X$ with additional perturbation on the loss value. The magnitude of the loss is controlled by varying $\varepsilon$. In comparison to the models from the literature (top row), our approach flattens the decision space significantly as demonstrated by the low loss values -- even for very large values of $\varepsilon$. Note that our proposed models (bottom row) were trained using $\alpha \sim B(2, 4)$ and $KL_{\lambda=10}$.}
        \label{fig:loss}
\end{figure*}

\begin{table*}[!t]
    \caption{Accuracy of different networks on the sets of original and adversarial images. All adversarial datasets are crafted using our \textit{gradient ensemble attack} approach with five unprotected VGGNet13, ResNet32 and ResNet50 models. For the last dataset (last column), we combined three VGG13, ResNet32 and ResNet50 models. The table shows the accuracy of three unprotected models on different adversarial datasets. 
    As one can see, multi-step adversaries do transfer between topologies. Further, our proposed method provides resilience against these transferable adversaries.}
    \label{tab:transfer}
    \centering
    \begin{small}
        \begin{tabular}{r|cc|cc|cc|cc|cc}
        &\multicolumn{2}{c|}{\bf Accuracy on} & \multicolumn{8}{c}{\bf Accuracy on Adversarial Dataset}\\
        &\multicolumn{2}{c|}{\bf Original Data} & \multicolumn{2}{c|}{\bf VGGNet13} & \multicolumn{2}{c|}{\bf ResNet32} & \multicolumn{2}{c|}{\bf ResNet50} & \multicolumn{2}{c}{\bf Combined} \\
        &Train &Val. & True& Adv. & True & Adv. & True & Adv. & True & Adv.\\
        \midrule \multicolumn{11}{l}{\textit{Base Models}}\\
        VGG13 \, & \, 91.2\% \, & \, \cellcolor{gray!25}\textbf{73.4}\% \, & \, 96.7\% \, & \, \cellcolor{gray!25}\textbf{0.9}\% \, & \, 96.2\% \, & \, 28.3\% \, & \, 96.9\% \, & \, 50.2\% \, & \, 98.4\% \, & \, \cellcolor{gray!25}\textbf{0.01}\% \\
        ResNet32 \, & \, 89.4\% \, & \, 66.7\% \, & \, 87.9\% \, & \, 67.6\% \, & \, 95.5\% \, & \, \cellcolor{gray!25}\textbf{16.0}\% \, & \, 95.3\% \, & \, 44.0\% \, & \, 96.1\% \, & \, \cellcolor{gray!25}\textbf{26.1}\% \\
        ResNet50 \, & \, 87.5\% \, & \, 62.8\% \, & \, 84.1\% \, & \, 64.9\% \, & \, 92.2\% \, & \, 24.6\% \, & \, 94.7\% \, & \, \cellcolor{gray!25}\textbf{30.9}\% \, & \, 94.5\% \, & \, \cellcolor{gray!25}\textbf{26.7}\% \\
        \midrule \multicolumn{11}{l}{Our Proposed Method with $\alpha \sim B(2, 4)$ and $KL_{\lambda=10}$}\\
        VGG13 \, & \, 90.6\% \, & \, \cellcolor{gray!25}\textbf{74.3}\% \, & \, 95.1\% \, & \, \cellcolor{gray!25}\textbf{53.7}\% \, & \, 96.6\% \, & \, 69.6\% \, & \, 97.1\% \, & \, 81.5\% \, & \, 98.1\% \, & \, \cellcolor{gray!25}\textbf{32.4}\% \\
        ResNet32 \, & \, 83.1\% \, & \, 59.8\% \, & \, 80.3\% \, & \, 67.8\% \, & \, 88.5\% \, & \, \cellcolor{gray!25}\textbf{38.5}\% \, & \, 89.5\% \, & \, 56.9\% \, & \, 90.2\% \, & \, \cellcolor{gray!25}\textbf{45.1}\% \\
        ResNet50 \, & \, 80.4\% \, & \, 62.4\% \, & \, 83.8\% \, & \, 70.3\% \, & \, 91.6\% \, & \, 44.1\% \, & \, 92.9\% \, & \, \cellcolor{gray!25}\textbf{56.4}\% \, & \, 93.6\% \, & \, \cellcolor{gray!25}\textbf{46.9}\% \\
        \end{tabular}
    \end{small}
\end{table*}

Figure \ref{fig:gradient} illustrates the loss of six different models based on the decision space of a single input image in two adversarial directions -- one taken from an unprotected model and the other one in direction of the related model. As depicted by Figure~\ref{fig:gradient}~(a), i.e., the image of the \textit{Base Model}, one can see an adversarial sub-space in close proximity of the input image as indicated by the high loss value (shown in red). Thus, it only takes a small change in  $\varepsilon$ ($\varepsilon=6$ is optimal in this particular case) to push the input image $X$ deep into this adversarial sub-space. A similar adversarial sub-space occurs in case of the VAT model as shown in Figure~\ref{fig:gradient}~(b). Here, the distance is greater, but the sub-space still exists. So, it takes a larger $\varepsilon$ to fool the VAT model.

In case of our proposed methods, using a fixed $\alpha$-value is not sufficient either as the blind-spot still exists, as illustrated in Figures~\ref{fig:gradient}~(d) and (e). Although our proposed methods with a fixed $\alpha$-value weaken the adversarial spot, they do not eliminate it. Probably, the $\alpha$-value is chosen too high in this case as there is another adversarial spot behind the first one. We found that sampling $\alpha \sim B(p,q)$ is essential to flatten the decision space. As demonstrated by Figure~\ref{fig:gradient}~(f) the decision space of our method with varying $\alpha$-values is significantly more flat than all the others. In fact, it is even more flat than the one of the PGD model -- especially for large $\varepsilon$-values as illustrated in Figures~\ref{fig:gradient}~(c) and (f), respectively. 

To verify these findings, we generated 128 adversarial images $X$ (using the FGSM) for each defense method and compared the loss values based on varying $\varepsilon\in[0,128]$ (c.f. Figure \ref{fig:loss}). As for the VAT model, adversarial attacks can be highly successful. In fact, $\varepsilon=26$ provides the greatest success-rate for single-step adversaries while for the \textit{Base Model} $\varepsilon=8$ works best (see Figures \ref{fig:loss}~(a) and (b)). This proves once again, that the adversarial spots are just a little further away compared to the \textit{Base Model}. This may be an explanation, on why the VAT model performs rather good on our generated adversaries. However, the VAT model is not substantially more robust against adversarial attacks than the unprotected base model -- it only requires a larger $\varepsilon$ to fool it.

In contrast, the PGD model and our proposed method clearly flatten the decision space and thereby strongly reduce the risk of adversarial spots (cf. Figure \ref{fig:loss}~(c) and (d)). As one can see, adversarial examples occur in greater distance compared to the \textit{Base Model} and occur significantly less frequent as the lower loss values demonstrate. However, there are adversarial sub-spaces which cause a high distraction of the PGD models. In contrast to the PGD model, our model (Virtual Alpha with $p=2$, $q=4$ and $\lambda=10$) provides a significantly more flattened space, i.e., high loss values rarely occur at all.  

Noticeably, for $\varepsilon<16$ the loss values of our model are significantly higher than the ones of the PGD model. In addition, loss values grow rapidly for small $\varepsilon$-values. The high average loss values for small $\varepsilon$-values may explain \textit{why} our model has a lower accuracy on our generated adversarial datasets than the PGD model as the loss values do not need do be maximal in order to indicate misclassification.

\section{Conclusion}
\label{sec:conclusion}
This work investigates the effects of adversarial attacks on deep learning networks. It analyzes different strategies for increasing a model's resilience and, thus, countervailing malicious attacks. The performance of the different defense strategies is compared across large sets of transferable, carefully generated adversaries. Next, three new approaches to improve resilience against such perturbations were first introduced and then compared against the state-of-the-art techniques \textit{Virtual Adversarial Training} and \textit{Projected Gradient Descent}. In addition, a novel adversarial attack method called \textit{gradient ensemble attack} has been introduced. Further, this work has demonstrated the transferability of adversaries, which have been crafted using our proposed method.

Within our investigations we have observed that VAT does not provide substantial resilience against adversarial perturbations as the adversarial sub-spaces are just pushed a little further away. However, the incidence of these spaces is similar to an unprotected model. Further, PGD trained models reduce the frequency of adversarial sub-spaces and strongly increase the distance to them. Yet, these sub-spaces still occur. Our proposed method, superimposing two images 
and minimizing the KL-Divergence between the two activations, reduces the risk of adversarial sub-spaces with high loss. In fact, our results demonstrate that these spaces rarely occur. However, the average loss value is significantly higher which explains why our models performed worse on our adversarial test sets. Nevertheless, our proposed method is very promising as it (i) is easily executable (it only requires an additional forward pass), and (ii) still provides a noticeable regularized decision space. 



Our ideas for future work are two-fold: (i) we will compare additional methods to further decrease the overall loss of our proposed method and thereby improve its performance on adversaries; (ii) we will investigate the effects of our \textit{gradient ensemble attack} for crafting strong and transferable adversaries in a wider context -- especially applying it to different white- and black-box attack scenarios.

\section*{Acknowledgment}

All three authors acknowledge support by the \href{https://www.ercis.org}{\emph{European Research Center for Information Systems (ERCIS)}}.







\bibliography{wcci2020}
\bibliographystyle{IEEEtran}
\end{document}